%% file: paper-727.tex
\pdfoutput=1
\documentclass[runningheads]{llncs}
\usepackage{graphicx}
\usepackage{xcolor}
\usepackage{multirow}
\usepackage{siunitx}
\usepackage{amsmath,amssymb,amsfonts}
\begin{document}
\title{Hyperparameter Tuning MLP's for Probabilistic Time Series Forecasting}
\titlerunning{Hyperparameter Tuning MLP's for TS Forecasting}
%
\author{Kiran Madhusudhanan$^*$ \orcidID{0000-0001-6356-8646} \and
Shayan Jawed$^*$ \orcidID{0009-0001-9130-8208} \and
Lars Schmidt-Thieme\orcidID{0000-0001-5729-6023}}
\authorrunning{K. Madhusudhanan et al.}
%
\institute{Information Systems and Machine Learning Lab \\
\& VWFS Data Analytics Research Center \\
University Of Hildesheim \\
Hildesheim, Germany \\
\email{\{madhusudhanan,shayan,schmidt-thieme\}@ismll.uni-hildesheim.de}
}
\maketitle              
\def\thefootnote{*}\footnotetext{These authors contributed equally to this work}
\def\thefootnote{\arabic{footnote}}
\begin{abstract}
Time series forecasting attempts to predict future events by analyzing past trends and patterns. Although well researched, certain critical aspects pertaining to the use of deep learning in time series forecasting remain ambiguous. Our research primarily focuses on examining the impact of specific hyperparameters related to time series, such as context length and validation strategy, on the performance of the state-of-the-art MLP model in time series forecasting. We have conducted a comprehensive series of experiments involving 4800 configurations per dataset across 20 time series forecasting datasets, and our findings demonstrate the importance of tuning these parameters. Furthermore, in this work, we introduce the largest metadataset for time series forecasting to date, named TSBench, comprising 97200 evaluations, which is a twentyfold increase compared to previous works in the field. Finally, we demonstrate the utility of the created metadataset on multi-fidelity hyperparameter optimization tasks.

\end{abstract}

\input{sections/introduction.tex}

\input{sections/related_works.tex}

\input{sections/preliminary.tex}

\input{sections/methodology.tex}

\input{sections/experiments.tex}

\input{sections/results.tex}

\input{sections/conclusion.tex}

\bibliography{reference}
\bibliographystyle{splncs04}

\end{document}

%% file: sections/introduction.tex
\section{Introduction}

 Time series forecasting is a machine learning technique that aims to capture historical patterns and use these patterns to predict the future values of the variables. 

 Time series datasets are generated by various physical phenomena that change over time and can be described by different mathematical equations or functions. Using deep learning techniques to model the underlying distribution is a frequent approach. However, choosing the optimal hyperparameters for a learning algorithm is a challenging problem in deep learning, known as hyperparameter optimization (HPO). HPO methods have been widely researched and applied in domains such as computer vision \cite{ullah2022meta}, and tabular datasets \cite{arango2hpo}, but they have received less attention within the time series domain. 

For instance, the context length, a crucial parameter that determines the extent of immediate history available to the model for forecasting, is often assigned a constant value across different datasets without careful tuning \cite{zhou2021informer,nie2022patchtst}. Despite evidence suggesting the importance of a context length as a hyperparameter \cite{Zeng2022AreTE,deng2023efficient}, subsequent research appears to overlook this parameter’s significance, maintaining a constant value across various datasets. While it could be argued that a longer context length is invariably beneficial, our experiments challenge this assumption. We demonstrate that the optimal context length is dependent on the dataset and varies according to the frequency and prediction horizon of the time series dataset.

Another point of contention within the time series forecasting community pertains to the use of validation splits. The validation split is typically chosen to best represent the test distribution. In most domains, a random sample of x\% of the training dataset can be used for validation, as the test dataset is usually an unknown random subset of the dataset during the training process. However, this is not the case for time series forecasting where, the test split is definitively the last few samples when the samples are arranged in chronological order, i.e., the test split is a time-wise split. 

While current works employ a time-wise validation split, the community lacks consensus on whether the forecasting model should be retrained on the validation split using the same hyperparameters, as done in \cite{nbeatsOreshkinCCB20}, or whether not to retrain on the validation split, as is common in other domains and in more recent time series forecasting papers such as \cite{Zeng2022AreTE,Autoformer021_bcc0d400,Madhusudhanan2021UNetIT}. This divergence underscores the need for further research and discussion within the community. In this paper, we rigorously benchmark across many configurations and datasets and analyze these specific validation split defining strategies that are applicable to time series forecasting.

Despite the general nature and extensive research in the field of time series forecasting, there is a noticeable lack of studies on hyperparameter optimization compared to other domains. In \cite{deng2023efficient}, the authors suggest the use of an AutoML framework that simultaneously optimizes the architecture and corresponding hyperparameters for a given dataset. However, this AutoML framework falls short in its ability to apply the knowledge gained from tuning one dataset to another.
A study more closely related to our work is \cite{Borchert2022MultiObjectiveMS}, where the authors compare various deep learning and classical models on 44 datasets and provide a metadataset consisting of evaluations and forecasts for all methods. However, this metadataset is limited to nearly five thousand runs, as reported in Table \ref{tbl:works}.
\begin{table}[t]
\centering
\caption{Summary statistics for evaluations considered in prior works.}
\resizebox{0.8\columnwidth}{!}{%
    \begin{tabular}{ccrcc}
        \textbf{Paper} &  \textbf{Venue} &  \textbf{\# HPs} & \textbf{\# Datasets} & \textbf{\# Evaluations}\\ \hline
        \cite{Borchert2022MultiObjectiveMS} &  ArXiv'22 & 107  & 44 & 4.7K\\ 
        \cite{deng2023efficient} &  ECML/PKDD'23 &200  & 24 & 4.8K\\
        \textbf{Our TSBench} &  - &4860  & 20 & 97 K\\ \hline
    \end{tabular}}
    \label{tbl:works}
\end{table}

Our work diverges from the aforementioned studies based on the number of evaluations performed. We evaluate 4860 configurations of the state-of-the-art muli-layer perceptron (MLP) model, named NLinear \cite{Zeng2022AreTE} for 20 datasets, creating the largest metadataset of 97200 evaluations for time series forecasting. In addition to the evaluations and forecasts, our work differs in the fact that we adapt the NLinear to probabilistic outputs while logging also the gradient statistics as in \cite{ZimLin2021a}, enabling the metadataset to be used for Learning Curve Forecasting techniques \cite{jawed2021multi}. This comprehensive approach sets our work apart in the field of time series forecasting. Concretely, we summarize the contributions of the paper as follows.

\begin{enumerate}

    \item We analyze the importance of time series specific hyperparameters like the validation strategy and context length for time series forecasting.
    \item We introduce \texttt{TSBench}, a benchmark for multi-fidelity optimization that provides probabilistic time series forecasts for 97200 hyperparameter evaluations on 20 datasets from the Monash Forecasting Repository \cite{godahewa2021monash}. 
    \item As a secondary task, we show the effectiveness of the TSBench dataset on multi-fidelity hyperparameter optimization.
\end{enumerate}

%% file: sections/related_works.tex
\section{Related Works}

The field of time series forecasting, particularly for long horizons, has recently garnered increased attention \cite{zhou2021informer,nie2022patchtst}. Numerous transformer models\cite{zhou2021informer,nie2022patchtst}, originally developed for the natural language processing (NLP) field where they have achieved significant results, have been adapted for this purpose. 
However, despite the sequential nature of both NLP and time series data, their distinct characteristics mean that directly applying transformer models from NLP to the time series domain may not yield optimal results. This is demonstrated in \cite{Zeng2022AreTE}, where a simple linear layer model with a specific normalization method outperforms transformer-based models on benchmark datasets. 
This finding is surprising given the relative simplicity of the linear model compared to the transformer models. 
In this work, we utilize the NLinear model proposed by \cite{Zeng2022AreTE} and assess its performance on the Monash Forecasting Repository \cite{godahewa2021monash}. The rapid execution time of the NLinear model enables us to test a larger number of configurations of a state-of-the-art model within a reasonable timeframe. This approach allows for a more comprehensive exploration of the model’s parameter space, potentially leading to improved performance and insights.

In the realm of increasingly expansive search spaces, the utilization of a metadataset with offline evaluations emerges as a technique that, while prevalent in other domains, appears to be rare in the literature pertaining to Time Series forecasting. The most widely adopted HPO method, Bayesian optimization, effectively maps a hyperparameter configuration to a corresponding validation loss, which is subsequently generated by the machine learning model. This technique was initially popularized by \cite{wistuba2015learning}, thereby paving the way for subsequent research into the application of transfer and non-transfer black box HPO \cite{volpp2019meta,feurer2015initializing}.
Metadatsets have also been extensively employed as a warm-start initialization technique \cite{feurer2015initializing} or for the purpose of transferring learning to the surrogate model \cite{wistuba2018scalable}. Furthermore, offline evaluations have been applied in the context of learning curve forecasting, where the objective is to predict the characteristics of a learning curve by embedding meta-knowledge that enables the model to exploit latent correlations among source dataset representations \cite{jawed2021multi}.

Despite the considerable body of research conducted on offline evaluations, the focus of these studies has predominantly been on simple regression and classification problems derived from the OpenML dataset \cite{vanschoren2014openml}, with a noticeable gap in the time series forecasting literature. To our knowledge, the work most closely aligned with ours is that of \cite{Borchert2022MultiObjectiveMS}, in which the authors evaluate 4.7K configurations across 44 datasets. However, our work diverges in three key respects: (1) we tune hyperparameters specific to time series, (2) we log gradient statistics at each epoch similar to \cite{ZimLin2021a} and (3) we provide evaluations that are 20 times more extensive in comparison to the aforementioned study.

%% file: sections/preliminary.tex
\section{Problem Statement}

Given a time series $(x, y) \sim q$ drawn
from an unknown random distribution $q$
where $x = (x_1, \ldots, x_C)$
and $y = (x_{C+1}, \ldots, x_{C+\delta})$
represent the observed history and future values,
respectively. Here, $x_i \in \mathbb{R}$
($D \in \mathbb{N}$) denotes the
observation made at relative time $i$. 
$x$ and $y$ are called univariate time series
if $D = 1$ and multivariate time series if $D > 1$.
Additionally, $C$ and $\delta$ represents here 
the Context Length and Forecast Horizon, respectively.

Univariate time series forecasting task attempts to
find a model $m: \mathbb{R}^C \to \mathbb{R}^\delta$
such that the expected loss 
$\ell: \mathbb{R}^\delta \times \mathbb{R}^\delta \to \mathbb{R}$ between the ground truth ($y$) and forecasted future values ($\hat{y}= m(x)$) is minimal:
$$
\mathop{\mathbb{E}}_{(x,y)\sim q} \ell(y, m(x))
$$

%% file: sections/methodology.tex
\section{MLPs for Time Series Forecasting}

\subsection{Nlinear Model}

In our research, we utilize a variant of linear MLP models, known as the NLinear \cite{Zeng2022AreTE}. This model employs an N-normalization technique to tackle the prevalent issue of distribution shift in time series datasets. The methodology involves subtracting the last value ($x_C$) of the sequence from the input ($x$), which is then passed through a linear layer to generate intermediate embedding ($z$). The subtracted values are subsequently added back to generate the forecasts ($\hat{y}$). 

\begin{align}
     z &= x - x_C \nonumber\\ 
     z &= \text{Linear}(z) \nonumber\\
     \hat{y} &= z + x_C
    \label{eq:nlinear}
\end{align}

The effectiveness of this approach has been demonstrated by \cite{Zeng2022AreTE} on seven benchmark datasets, thereby questioning the efficacy of current transformer-based models for time series forecasting. The study also claims that deeper models fail to enhance performance of the Linear model on the seven benchmark datasets. This raises an interesting question : Can linear models outperform their deeper non-linear counterparts in general?.

\begin{figure}[t]
    \centering
    \includegraphics[width=0.8\linewidth]{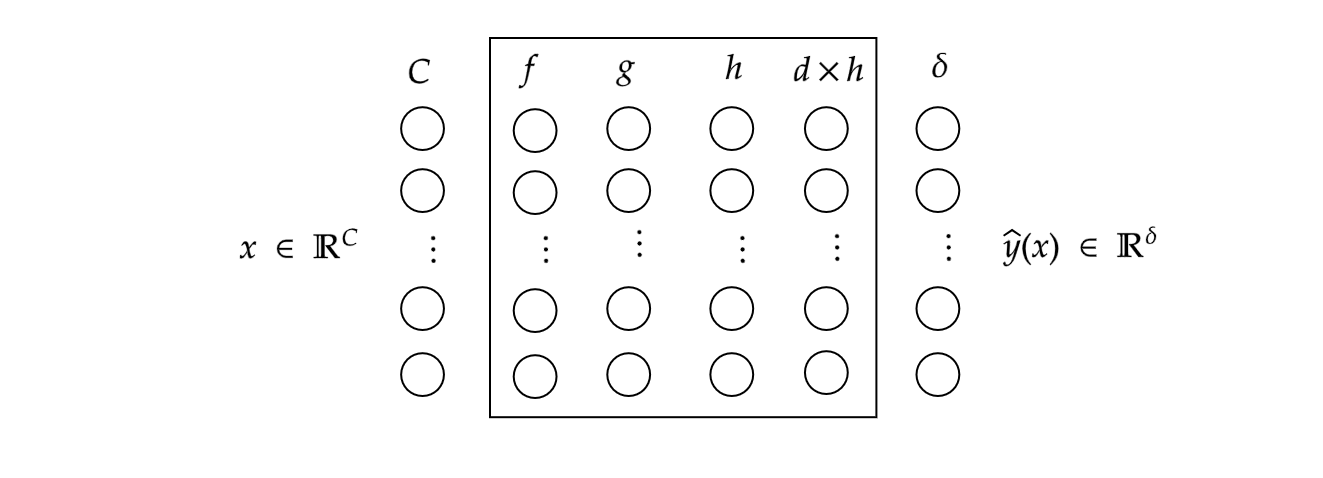}
    \caption{\textbf{Model Architecture}. The parameters $C$ and $\delta$ are indicative of the context length and forecast horizon, respectively. The hidden layers within the model are represented by $f$, $g$, and $h$, and are interspersed with ELU non-linearity. The parameter $d$ signifies the distribution parameters that is learned per prediction time step. }
    \label{fig:archi}
\end{figure}

\section{Hyperparameters}

To address the previous question, we construct deeper model architecture variants in line with the methodology outlined by \cite{jomaa2021dataset2vec}, by augmenting the NLinear model with three submodules, ($f \circ g \circ h$), each implemented as a fully connected layer with ELU \cite{clevert2015fast} activation function as in Figure \ref{fig:archi}.

\begin{enumerate}
    \setcounter{enumi}{0}
    \item \textbf{Model Architectures}
    \begin{itemize}
        \item \texttt{Base}: We choose the NLinear model as proposed in \cite{Zeng2022AreTE}, a single layer network without any non-linearity as the base model. For instance, the structure could be represented as $C$-$\delta$, where $C$ denotes the Context length (Input layer) and $\delta$ (Output layer) signifies the forecast horizon.
        \item \texttt{Diamond}: The depth of the NLinear model is augmented by incorporating hidden layers that feature an expanding layer at the center, separated by ELU nonlinearity, thereby adopting a diamond-like shape. The structure could $C$-$f$-$g$-$h$-$\delta$ be represented as $C$-32-64-32-$\delta$.
        \item \texttt{Contracting}: The hidden layers of the model are designed with decreasing number of neurons at each successive layer. eg: C-128-64-32-$\delta$.
        \item \texttt{Square}: The hidden layers have the same number of neurons at each hidden layer. eg: C-64-64-64-$\delta$.
        \item \texttt{Funnel}: The inverse of the \texttt{Diamond} architecture with a contracting layer at the middle of the network, eg: C-64-32-64-$\delta$
        \item \texttt{Expanding}: The hidden layers of the model are designed with decreasing number of neurons at each successive layer. eg: C-32-64-128-$\delta$
    \end{itemize}

\end{enumerate}

The MLP architectures delineated previously have been adapted to leverage the probabilistic forecasting capabilities provided by the GluonTS framework \cite{alexandrov2019gluonts}. This is achieved by predicting the parameters of distribution $d$ at each predictive timestep. For instance, under the assumption of a Gaussian distribution, the model is tasked with learning the mean and variance as distribution parameters for each timestep. The aforementioned MLP architectures can be extended to generate probabilistic forecasts by incorporating a distribution layer atop the model, as depicted in Figure \ref{fig:archi}. In our experimental setup, we utilize the Student-T distribution, which is the default distribution in GluonTS, and we experiment with varying the number of hidden parameters in the distribution layer.

\begin{enumerate}
    \setcounter{enumi}{1}
        \item \textbf{Distribution hidden layer $(d)$} :  The selection of the distribution hidden layer is made from a grid that includes the values 1, 2, and 10. 
        Default size in GluonTS is 1, however the size is increased to 2 and 10 as increase in layer size have been observed to enhance the performance of the model. 
\end{enumerate}

\subsection{Time Series Specific Configuration}
\label{sec:config}

The \textbf{context length}, also known as the lookback window, is a critical parameter in time series forecasting. It determines the amount of historical data used to predict future values of a time series.
In their study \cite{Zeng2022AreTE}, the authors examined the effects of context lengths on long-term forecasting. They found that the performance of most transformer-based methods deteriorates with an increase in context length. Similarly, \cite{deng2023efficient} compared the importance of various parameters used in time series forecasting and found that context length is one of the most important hyperparameters to tune.
Despite these findings, recent literature on time series forecasting often uses a constant context length across multiple forecasting tasks \cite{Autoformer021_bcc0d400,zhou2021informer,nie2022patchtst}. Our work aligns with these previous studies in analyzing the importance of context length. However, unlike \cite{deng2023efficient}, who consider context length as multiples of seasonality, we do not make any assumptions about the context length for a dataset. Our grid of context length ranges from very short to very long across all datasets.

\begin{enumerate}
    \setcounter{enumi}{2}
    \item \textbf{Context Length (C)}: 

    In our study, the specific values we have selected for our grid are {2, 7, 24, 100, 300}. These values represent a summary of the default context lengths provided in the Monash Forecasting Repository.

\end{enumerate}

\textbf{Validation Strategy} for time series forecasting is not standardized in the literature. For instance, a well-known forecasting model NBEATS \cite{nbeatsOreshkinCCB20} uses a validation set to select the hyperparameters and then re-trains the model on the combined training and validation sets, while some recent forecasting models such as NLinear \cite{Zeng2022AreTE} skip the re-training step altogether. Therefore, we consider the validation strategy as another hyperparameter to be optimized. 

\begin{enumerate}
    \setcounter{enumi}{3}
    \item \textbf{Validation Strategy}:
    \begin{itemize}
        \item \texttt{Out-of-Sample} (OOS): The validation split is a time-wise split replicating the test split, however the model is not retrained.
        \item \texttt{Retrain-Out-Of-Sample} (Re-OOS): The validation split is a time-wise split replicating the test split, but the model is  retrained on the validation split using the hyperparameters chosen from the validation split.
        \item \texttt{In-Sample} (IS): The validation split is randomly sampled from the dataset as in \cite{salinas2020deepar}. 
    \end{itemize}
\end{enumerate}

\subsection{Training Specific Configurations}

In our comprehensive evaluation, we take into account not only the specificities of time series and model configurations, but also the configurations of model training hyperparameters. To this extent we tune two training hyperparamters namely the learning rate and the weight decay. For the learning rate, we choose from the possible options of 0.01, 0.001 or 0.0001, while retaining the default value of 0.001 in the selection. To our understanding, previous studies have not considered regularized time series forecasting evaluations \cite{Borchert2022MultiObjectiveMS}, \cite{deng2023efficient}. The significance of effective regularization for neural network models in the tabular domain has been emphasized in recent research \cite{kadra2021well}. Consequently, we have incorporated the option for our model to apply regularization through weight decay.
Finally, we repeat each experiment 3 times for consistency and report the standard deviation. The configurations are summarized in Table \ref{tbl:config}.

\begin{table}[t]
\centering
\caption{Summary of configurations used for generating the \texttt{TSBench} metadataset.}
\resizebox{0.9\columnwidth}{!}{%
\label{tbl:config}
\begin{tabular}{|l|l|l|}
\hline
\textbf{Configuration}                & \textbf{Hyperparameter}            & \textbf{Values}                                                      \\
\hline
\multirow{2}{*}{Time Series} & Context Length            & {[}2, 7, 24, 100, 300{]}                                    \\
                             & Validation Strategy       & {[}OOS, Re-OOS{]}                                           \\
\hline
\multirow{2}{*}{Model}       & Architecture              & \begin{tabular}[c]{@{}l@{}}{[}Base, Diamond, Contracting, \\ Square, Funnel, Expanding{]}\end{tabular} \\
                             & Distribution Hidden State & {[}1, 2, 10{]}                                              \\
\hline
\multirow{3}{*}{Training}    & Learning Rate             & {[}0.01, 0.001, 0.0001{]}                                   \\
                             & Weight Decay              & {[}0, 0.1, 0.5{]}                                           \\
                             & Seeds                     & {[}100, 101, 102{]}                                        \\
\hline
\end{tabular}}
\end{table}

\subsection{TSBench-Metadataset}

Metadatasets have shown to improve the performance of hyperparameter optimization \cite{jomaa2021dataset2vec,arango2hpo} and often provide a qualitative basis to focus efforts in both manual algorithm design and automated hyperparameter optimization. In this paper, we follow the notable metadataset work by \cite{shah2021autoai}, where the authors record arguably the most important metafeatures required for the secondary tasks like learning curve forecasting \cite{jawed2021multi} or transfer hyperparameter optimization \cite{arango2hpo}. We evaluate 4800 configurations per dataset for 20 datasets, each evaluated for 50 epochs, and log the results as our \texttt{TSBench} metadaset.

In our work, we collect metafeatures for each run at two levels of granularity. Firstly, we log metafeatures per configuration on a coarse scale, and secondly, on a fine-grained scale per epoch. At the coarse scale, we log the basic hyperparameter configurations mentioned in the previous sections and additionally include features such as the number of trainable parameters and dataset metafeatures like time series data frequency, seasonality, etc. At the fine-grained per epoch scale, we capture the train, validation, test losses and various metrics reported per epoch. This allows the user of the metadataset to perform hyperparameter optimizations based on metrics other than the train loss. GluonTS \cite{alexandrov2019gluonts} offers numerous probabilistic and point-wise evaluation metrics for time series forecasting including the Quantile Loss, CRPS, seasonal error to name a few. In addition, learning curve forecasting methods like \cite{jawed2021multi} could benefit from layer-wise gradient statistics such as the max, mean and quantiles as additional covariates to predict the trajectory of a particular hyperparameter run. \texttt{TSBench} also logs this information along with learning rate and runtime information. Table \ref{tbl:metrics} provides an overview of all the logged metadataset features.

\begin{table}[t]
\caption{Summary of metrics logged for generating the \texttt{TSBench} metadataset.}
\label{tbl:metrics}
\resizebox{\columnwidth}{!}{%
\begin{tabular}{|l|l|l|}
\hline
\textbf{Granularity}                    & \textbf{Metric}               & \textbf{Value}                                                                                                                                               \\ \hline
\multirow{5}{*}{Epoch}         & Losses               & Negative log likelihood loss                                                                                                                        \\ \cline{2-3} 
                               & Metrics              & \begin{tabular}[c]{@{}l@{}}MSE, MASE, MAPE, QuantileLoss  at quantile interval of 10,\\ RMSE, NRMSE, ND, MAE and weighted QuantileLoss\end{tabular} \\ \cline{2-3} 
                               & Layer-wise Gradients & Max, Mean, Median, Std, and quantiles at intercal of 10                                                                                             \\ \cline{2-3} 
                               & Learning Rate        & -                                                                                                                                                   \\ \cline{2-3} 
                               & Runtine              & -                                                                                                                                                   \\ \hline
\multirow{2}{*}{Configuration} & Architecture         & \begin{tabular}[c]{@{}l@{}}Activations, Model Architecture, Hyperparameters,\\ Model Complexity,\end{tabular}                                       \\ \cline{2-3} 
                               & Dataset Features     & Context Length, Prediction Length, Seasonality, Frequency                                                                                           \\ \hline
\end{tabular}}
\end{table}

%% file: sections/experiments.tex
\section{Experimental Setup}

\textbf{Evaluation:} All models were trained using the negative log likelihood loss to generate probabilistic forecasts, and CRPS score \cite{rasul2021multivariate} was reported as an uncertainty error metric in the supplementary material. However, in order to compare with the Monash Forecasting Repository results \cite{godahewa2021monash}, we evaluate using MASE error metric.
\begin{align}
    \text{MASE}&=\frac{1}{\delta}\sum_{j=0}^{\delta} \frac{|y_j - \hat{y_j}|}{|y_j - \hat{y_j}^{\text{Naive}}|}
\end{align}
\textbf{Data:} Monash Forecasting Repository \cite{godahewa2021monash}, which is a collection of 50 datasets that are derived from 26 original real-world datasets by sampling time series data at different frequencies. We randomly select 20 datasets from this collection that do not have missing values and have varying characteristics, such as length, number of series, etc., to capture the diversity of real-world time series data.

\hspace{-1.5em}\textbf{Framework:} In our experiments, we set a batch size of 64 and a number of batches per epoch of 50 for all runs. The models were trained for a total of 50 epochs, and the model with the lowest validation loss was selected for evaluation on the test set. 
For the \texttt{Re-OOS} validation strategy, since the model was trained using an average loss instead of sum of losses, retraining the model with the validation split (\texttt{Re-OOS}) should have only a negligible impact on the chosen hyperparameters. \texttt{In-Sample} validation implementation was not straight forward in GluonTS, as the framework expects the splits to be in continuous time and limiting our study to \texttt{Re-OOS} and \texttt{OOS}.
All experiments were conducted on Intel E5-2670v2 CPU cores using Pytorch 1.12, Pytorch-lightning 1.6.5, and MXNet 1.9. The training process took approximately one month on 40 nodes with the same CPU configuration. Code\footnote{https://github.com/18kiran12/TSBench.git} is made public and can be easily reused to generate a larger metadataset with other state-of-the-art models from the GluonTS\cite{alexandrov2019gluonts} framework, including NBEATS \cite{nbeatsOreshkinCCB20}, DeepAR \cite{salinas2020deepar}, among others. For HPO methods, we use the SMAC3 framework \cite{lindauerjmlr22aSMAC}.

%% file: sections/results.tex
\section{Results}
\begin{table}[t]
\centering
\caption{Comparison of \texttt{TSBench} results with the best performing model (TBATS) from Monash Benchmark reported on the MASE error. Best results are marked in bold. Standard deviations over multiple runs are indicated in brackets. We also provide the best overall result on the dataset across different models for reference.} 
\resizebox{\columnwidth}{!}{%
\begin{tabular}{l|cc|c|c}
\hline
Datasets           & Train                           & Retrain                         & \multicolumn{1}{l}{\begin{tabular}[c]{@{}l@{}}Monash - TBATS \end{tabular}}   & \multicolumn{1}{|l}{\begin{tabular}[c]{@{}l@{}}Monash - Best Overall \end{tabular}}\\
\hline
Aus. Elecdemand   & 1.693 (0.133)          & 1.667  (0.199)                                              & \textbf{1.174}        & 0.705       \\
Bitcoin           & 7.370 (2.207)           & 8.351 (1.30)                                               & \textbf{4.611}        & 2.664       \\
FRED-MD           & 0.569 (0.019)          & 0.580 (0.029)                                              & \textbf{0.502}        & 0.468       \\
Hospital          & 0.787 (0.020)              & 0.794 (0.025)                                               & \textbf{0.768}        & 0.761       \\
KDD               & 1.172 (0.011)          & \textbf{1.131 (0.016)}                                       & 1.394                 & 1.185                \\
M1 Monthly        & 1.565 (0.017)          & 1.577 (0.030)                                               & \textbf{1.118}        & 1.074       \\
M1 Quarterly      & 2.360 (0.239)           & 2.511 (0.283)                                               & \textbf{1.694}        & 1.658       \\
M1 Yearly         & 4.537 (0.028)          & 4.393 (0.018)                                               & \textbf{3.499}        & 3.499       \\
M3 Monthly        & 1.143 (0.006)          & 1.137  (0.011)                                              & \textbf{0.861}        & 0.861       \\
M3 Quarterly      & 1.327 (0.025)          & 1.313 (0.031)                                               & \textbf{1.256}        & 1.117       \\
M3 Yearly         & 4.168 (0.063)          & 35.640 (33.024)                                              & \textbf{3.127}        & 2.774       \\
M4 Hourly         & 1.421 (0.192)          & \textbf{1.188 (0.114)}                                       & 2.663                 & 1.662                \\
M4 Weekly         & 0.438 (0.014)          & \textbf{0.434 (0.002)}                                       & 0.504                 & 0.453                \\
NN5 Daily         & \textbf{0.800 (0.004)}   & 0.803 (0.007)                                               & 0.858                 & 0.858                \\
NN5 Weekly        & 0.875 (0.063)          & 1.143  (0.050)                                            & \textbf{0.872}        & 0.808       \\
Tourism Monthly   & \textbf{1.603 (0.002)} & 1.628 (0.077)                                               & 1.751          &  1.409      \\
Tourism Quarterly & 1.731 (0.018)          & \textbf{1.631 (0.038)}                                       & 1.835           & 1.475       \\
Tourism Yearly    & 6.283 (1.816)          & 5.744  (2.22)                                              & \textbf{3.685}        & 2.977       \\
Traffic Hourly    & 0.802 (0.002)          & \textbf{0.801 (0.013)}                                       & 2.482                 & 0.821                \\
Traffic Weekly    & 1.187 (0.006)          & 1.187 (0.010)                                               & \textbf{1.148}        & 1.094                                                                                              \\
\hline
\end{tabular}}
\label{tbl:resultsmonash}
\end{table}

In Table \ref{tbl:resultsmonash}, we present a comparison between the best performing NLinear and the best performing model TBATS from the Monash forecasting repository results. It is important to note that the TBATS model was able to outperform several powerful models, including NBEATS, DeepAR, and Transformer models across the different datasets, making it a really strong baseline to outperform. 

\subsubsection*{RQ1 How does NLinear compare to Monash's best model?}

The NLinear model outperforms the best model TBATS from the Monash Forecasting repository on 7 out of 20 datasets of varying granularities. Additionally, when compared to the best overall results from the Monash, the NLinear model performs well on 5 out of 20 datasets. Given that a data-specific baseline is a challenging benchmark to surpass, the NLinear model performs on par with, if not better than, other deep learning baselines from the Monash forecasting results. This underscores the importance of considering carefully tuned linear models as a baseline for time series forecasting.

\subsubsection*{RQ2: Should we retrain on the validation data?}

One of the contributions of this work is to address the uncertainty regarding whether deep learning models need to be retrained on validation data. Our results, presented in Table \ref{tbl:resultsmonash}, indicate that \texttt{Re-OOS} offers only a slight advantage compared to \texttt{OOS}. This is consistent with current trends in time series forecasting, where retraining on validation data is ignored. We reason that this may be due to suboptimal hyperparameter fitting, as the addition of validation data may require changes to the hyperparameters selected on the training data. 

It is of significant importance to highlight that a considerable decline in the performance of the M3 yearly dataset was observed after the retraining process. This shows an extreme scenario where the retraining process adversely impacted the dataset's performance.

\subsubsection*{RQ3: What is a useful Context Length?}

\begin{figure}[t]
    \begin{minipage}[b]{0.45\linewidth}
        \centering
        \includegraphics[width=\linewidth]{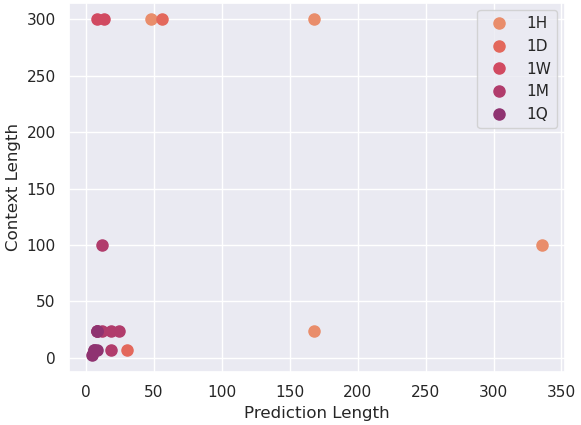}
        \caption{Prediction length vs Context Length colored by Frequency of dataset. Longer Prediction length.}
        \label{fig:ContextLen}
    \end{minipage}
    \hspace{0.5cm}
    \begin{minipage}[b]{0.45\linewidth}
        \centering
        \includegraphics[width=\linewidth]{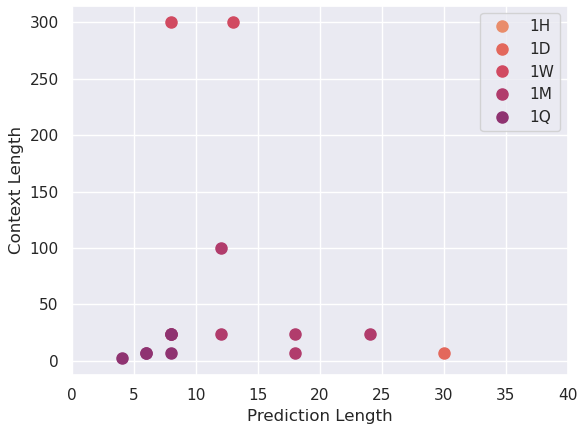}
        \caption{Prediction length vs Context Length colored by Frequency of dataset. Shorter Prediction length.}
        \label{fig:freqContext}
    \end{minipage}
\end{figure}

The length of context required by a model is influenced by several factors. Our analysis suggests that the context length is a function of both the frequency of the dataset and the length of the forecast horizon. When given the option to select from a range of model complexities and context lengths, the model often chose longer context lengths for longer forecast horizons, as shown in Figure \ref{fig:ContextLen} and \ref{fig:freqContext}. Additionally, our findings indicate that the frequency of the dataset also impacts the chosen context length. For instance, datasets with hourly (1H) and daily (1D) frequencies are more likely to have longer context lengths than those with yearly (1Y) or quarterly (1Q) frequencies. Figure \ref{fig:ContextLen} depicts the correlation between the length of the context and the prediction horizon, as well as the frequency of a specific time series dataset. Generally, a more extensive context is advantageous for the model when a larger forecast horizon is required. However, this pattern diminishes for monthly, yearly, and quarterly datasets. In these instances, the model performs satisfactorily with a shorter context length.

\subsubsection*{RQ4: What hyperparameters are Important?}

\begin{figure}[t]
    \begin{minipage}[b]{0.47\linewidth}
        \centering
        \includegraphics[width=\linewidth]{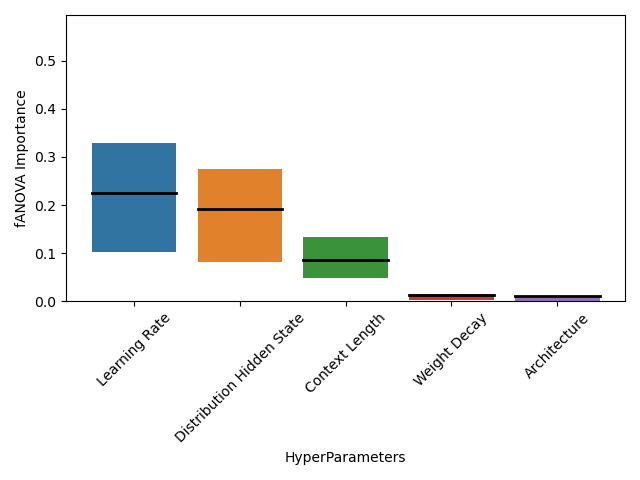}
        \caption{Hyperparameter importance score}
        \label{fig:importance}
    \end{minipage}
    \hspace{0.5cm}
    \begin{minipage}[b]{0.47\linewidth}
        \centering
        \includegraphics[width=\linewidth]{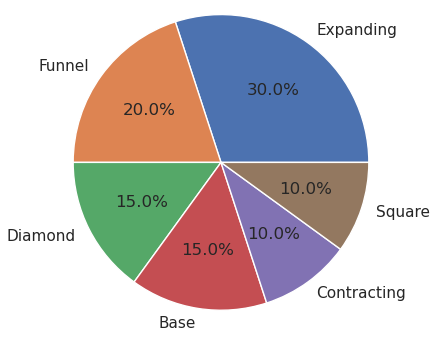}
        \caption{Architecture selection globally across multiple datasets}
        \label{fig:architectureSelection}
    \end{minipage}
\end{figure}

To evaluate the significance of various hyperparameters in forecasting, we utilized an fANOVA test as outlined in \cite{deng2023efficient}. This test employs a random forest model to capture the relationship between the hyperparameters and forecast accuracy, using the hyperparameters as input. A functional ANOVA is then applied to determine the importance of each hyperparameter. The results are depicted in Figure \ref{fig:importance}. The initial learning rate selected appears to have a substantial impact on model performance, even when the model is configured with the Adam optimizer. Additionally, the hidden layer used to learn the probabilistic distribution of the forecast is also an important parameter. And most importantly, context length has a significant effect on model accuracy and needs to be carefully tuned per dataset.

\subsubsection*{RQ4: Can linear models outperform non-linear models?}

In this study, we allowed the model to experiment with deeper architectures and ELU activations to determine whether a linear model consistently outperforms deeper models with various architectures, as described in Section \ref{sec:config}. Our findings, presented in Figure \ref{fig:architectureSelection}, indicate that deeper models can indeed be useful, however, considering the hyperparameter importance of architecture in Figure \ref{fig:importance}, a Linear MLP is a strong baseline in most cases.

\subsubsection*{RQ5: Can TSBench be effectively used for HPO?}
\begin{figure}[]
    \centering
    \includegraphics[width=0.7\linewidth]{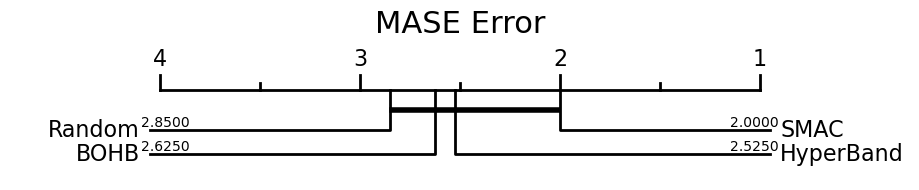}
    \caption{Critical Difference Diagram Rank@50}
    \label{fig:cd}
\end{figure}

The utilization of HPO as a supplementary meta-task to demonstrate the efficacy of the constructed metadataset is a prevalent approach \cite{arango2hpo}. In this study, we applied four distinct HPO techniques to the TSBench dataset. We start with a rudimentary baseline that randomly selects a single hyperparameter from a pool of 50 trials. Secondly, we employed the HyperBand \cite{li2017hyperband}, a bandit-based HPO approach that conducts multiple successive halving operations to identify optimal configurations.
Further, we utilized model-based algorithms such as SMAC \cite{lindauerjmlr22aSMAC} and BOHB, which utilize surrogate models to select the most promising hyperparameter evaluations. Specifically, SMAC employs a random forest model as its surrogate, while BOHB \cite{falkner2018bohb} adopts a Bayesian optimization algorithm. Each algorithm was permitted a total of 50 trials per dataset to select the best hyperparameters. The outcomes, presented in Figure \ref{fig:cd} across the 20 datasets in a critical difference diagram \cite{arango2hpo}, reveal that SMAC, HyperBand and BOHB outperform the Random strategy showing the effectiveness of the \texttt{TSBench} for HPO.

%% file: sections/conclusion.tex
\section{Conclusion}

Clarity regarding the significance of hyperparameters and the choice of validation strategy in time series forecasting literature is often lacking. This study aims to address these ambiguities by assessing the performance of the state-of-the-art MLP model on 20 univariate datasets.
Our findings highlight the importance of tuning the context length for time series forecasting tasks and treating the validation strategy as a hyperparameter. Interestingly, while deeper MLP models may offer performance enhancements on certain datasets, our results affirm the robustness of a linear MLP model as a formidable baseline.
Furthermore, we introduce TSBench, an extensive metadataset for time series forecasting to date, and demonstrate its efficacy in HPO tasks.

%% file: paper-727.bbl
\begin{thebibliography}{10}
\providecommand{\url}[1]{\texttt{#1}}
\providecommand{\urlprefix}{URL }
\providecommand{\doi}[1]{https://doi.org/#1}

\bibitem{alexandrov2019gluonts}
Alexandrov, A., Benidis, K., Bohlke-Schneider, M., Flunkert, V., Gasthaus, J.,
  Januschowski, T., Maddix, D.C., Rangapuram, S., Salinas, D., Schulz, J.,
  et~al.: Gluonts: Probabilistic time series models in python. ArXiv  (2019)

\bibitem{arango2hpo}
Arango, S.P., Jomaa, H.S., Wistuba, M., Grabocka, J.: Hpo-b: A large-scale
  reproducible benchmark for black-box hpo based on openml. In: NeurIPS
  Datasets and Benchmarks Track (2021)

\bibitem{Borchert2022MultiObjectiveMS}
Borchert, O., Salinas, D., Flunkert, V., Januschowski, T., Gunnemann, S.:
  Multi-objective model selection for time series forecasting. ArXiv  (2022)

\bibitem{clevert2015fast}
Clevert, D.A., Unterthiner, T., Hochreiter, S.: Fast and accurate deep network
  learning by exponential linear units (elus). ICLR  (2015)

\bibitem{deng2023efficient}
Deng, D., Karl, F., Hutter, F., Bischl, B., Lindauer, M.: Efficient automated
  deep learning for time series forecasting. In: ECML PKDD. pp. 664--680.
  Springer (2023)

\bibitem{falkner2018bohb}
Falkner, S., Klein, A., Hutter, F.: Bohb: Robust and efficient hyperparameter
  optimization at scale. In: ICML. pp. 1437--1446. PMLR (2018)

\bibitem{feurer2015initializing}
Feurer, M., Springenberg, J., Hutter, F.: Initializing bayesian hyperparameter
  optimization via meta-learning. In: AAAI. vol.~29 (2015)

\bibitem{godahewa2021monash}
Godahewa, R., Bergmeir, C., Webb, G.I., Hyndman, R.J., Montero-Manso, P.:
  Monash time series forecasting archive. NeurIPS Datasets and Benchmarks
  (2021)

\bibitem{jawed2021multi}
Jawed, S., Jomaa, H., Schmidt-Thieme, L., Grabocka, J.: Multi-task learning
  curve forecasting across hyperparameter configurations and datasets. In: ECML
  PKDD. pp. 485--501 (2021)

\bibitem{jomaa2021dataset2vec}
Jomaa, H.S., Schmidt-Thieme, L., Grabocka, J.: Dataset2vec: Learning dataset
  meta-features. Data Mining and Knowledge Discovery  \textbf{35},  964--985
  (2021)

\bibitem{kadra2021well}
Kadra, A., Lindauer, M., Hutter, F., Grabocka, J.: Well-tuned simple nets excel
  on tabular datasets. NeurIPS  \textbf{34},  23928--23941 (2021)

\bibitem{li2017hyperband}
Li, L., Jamieson, K., DeSalvo, G., Rostamizadeh, A., Talwalkar, A.: Hyperband:
  A novel bandit-based approach to hyperparameter optimization. JMLR
  \textbf{18}(1) (2017)

\bibitem{lindauerjmlr22aSMAC}
Lindauer, M., Eggensperger, K., Feurer, M., Biedenkapp, A., Deng, D.,
  Benjamins, C., Ruhkopf, T., Sass, R., Hutter, F.: Smac3: A versatile bayesian
  optimization package for hyperparameter optimization. JMLR  \textbf{23}(54),
  ~1--9 (2022)

\bibitem{Madhusudhanan2021UNetIT}
Madhusudhanan, K., Burchert, J., Duong-Trung, N., Born, S., Schmidt-Thieme, L.:
  U-net inspired transformer architecture for far horizon time series
  forecasting. In: ECML/PKDD (2021)

\bibitem{nie2022patchtst}
Nie, Y., Nguyen, N.H., Sinthong, P., Kalagnanam, J.: A time series is worth 64
  words: Long-term forecasting with transformers. ICLR  (2023)

\bibitem{nbeatsOreshkinCCB20}
Oreshkin, B.N., Carpov, D., Chapados, N., Bengio, Y.: {N-BEATS:} neural basis
  expansion analysis for interpretable time series forecasting. In: ICLR (2020)

\bibitem{rasul2021multivariate}
Rasul, K., Sheikh, A.S., Schuster, I., Bergmann, U.M., Vollgraf, R.:
  Multivariate probabilistic time series forecasting via conditioned
  normalizing flows. In: ICLR (2021)

\bibitem{salinas2020deepar}
Salinas, D., Flunkert, V., Gasthaus, J., Januschowski, T.: Deepar:
  Probabilistic forecasting with autoregressive recurrent networks. JMLR
  \textbf{36}(3),  1181--1191 (2020)

\bibitem{shah2021autoai}
Shah, S.Y., Patel, D., Vu, L., Dang, X.H., Chen, B., Kirchner, P., Samulowitz,
  H., Wood, D., Bramble, G., Gifford, W.M., et~al.: Autoai-ts: Autoai for time
  series forecasting. In: SIGMOD. pp. 2584--2596 (2021)

\bibitem{ullah2022meta}
Ullah, I., Carri{\'o}n-Ojeda, D., Escalera, S., Guyon, I., Huisman, M., Mohr,
  F., van Rijn, J.N., Sun, H., Vanschoren, J., Vu, P.A.: Meta-album:
  Multi-domain meta-dataset for few-shot image classification. NeurIPS
  \textbf{35},  3232--3247 (2022)

\bibitem{vanschoren2014openml}
Vanschoren, J., Van~Rijn, J.N., Bischl, B., Torgo, L.: Openml: networked
  science in machine learning. ACM SIGKDD Explorations Newsletter
  \textbf{15}(2),  49--60 (2014)

\bibitem{volpp2019meta}
Volpp, M., Fr{\"o}hlich, L.P., Fischer, K., Doerr, A., Falkner, S., Hutter, F.,
  Daniel, C.: Meta-learning acquisition functions for transfer learning in
  bayesian optimization. ICLR  (2020)

\bibitem{wistuba2015learning}
Wistuba, M., Schilling, N., Schmidt-Thieme, L.: Learning hyperparameter
  optimization initializations. In: DSAA. pp. 1--10. IEEE (2015)

\bibitem{wistuba2018scalable}
Wistuba, M., Schilling, N., Schmidt-Thieme, L.: Scalable gaussian process-based
  transfer surrogates for hyperparameter optimization. Machine Learning
  \textbf{107}(1),  43--78 (2018)

\bibitem{Autoformer021_bcc0d400}
Wu, H., Xu, J., Wang, J., Long, M.: Autoformer: Decomposition transformers with
  auto-correlation for long-term series forecasting. In: NeurIPS. vol.~34, pp.
  22419--22430 (2021)

\bibitem{Zeng2022AreTE}
Zeng, A., Chen, M., Zhang, L., Xu, Q.: Are transformers effective for time
  series forecasting? AAAI  (2023)

\bibitem{zhou2021informer}
Zhou, H., Zhang, S., Peng, J., Zhang, S., Li, J., Xiong, H., Zhang, W.:
  Informer: Beyond efficient transformer for long sequence time-series
  forecasting. In: AAAI. vol.~35, pp. 11106--11115 (2021)

\bibitem{ZimLin2021a}
Zimmer, L., Lindauer, M., Hutter, F.: Auto-pytorch tabular: Multi-fidelity
  metalearning for efficient and robust autodl. IEEE Transactions on Pattern
  Analysis and Machine Intelligence  \textbf{43}(9),  3079 -- 3090 (2021)

\end{thebibliography}
